\documentclass[12pt,a4paper]{article}
\pdfoutput=1
\usepackage{graphicx}
\usepackage{cite}
\usepackage{amssymb}
\usepackage{physics}
\usepackage{floatpag}
\floatpagestyle{empty}
\begin{document}
	\title{Real-Time Style Transfer With Strength Control}
	%
	%
	\author{Victor Kitov\\
		\emph{Lomonosov Moscow State University, Moscow, Russia} \\
        \emph{Plekhanov Russian University of Economics, Moscow, Russia}  \\
		v.v.kitov@yandex.ru
	}
			
	\maketitle	
	
	\begin{figure}
		\includegraphics[width=\textwidth]{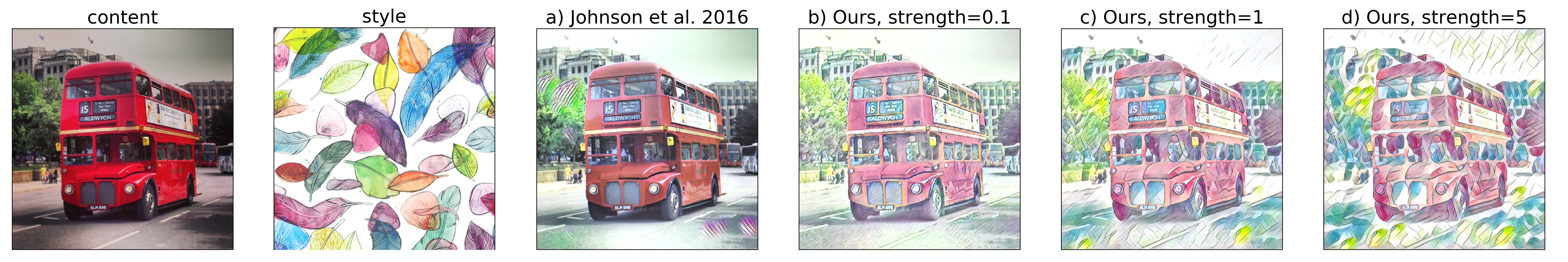}
		\caption{Style transfer method proposed in this article omits artifacts generated by Johnson et al.~\cite{johnson2016perceptual} (a) and allows continuous control of stylization strength (b)-(d).} \label{fig:headline}
	\end{figure}

\begin{abstract}
\emph{Style transfer} is a problem of rendering a content image in the style of another style image. A natural and common practical task in applications of style transfer is to adjust the strength of stylization. Algorithm of Gatys et al.~\cite{gatys2016image} provides this ability by changing the weighting factors of content and style losses but is computationally inefficient. \emph{Real-time style transfer} introduced by Johnson et al.~\cite{johnson2016perceptual} enables fast stylization of any image by passing it through a pre-trained transformer network. Although fast, this architecture is not able to continuously adjust style strength. We propose an extension to real-time style transfer that allows direct control of style strength at inference, still requiring only a single transformer network. We conduct qualitative and quantitative experiments that demonstrate that the proposed method is capable of smooth stylization strength control and removes certain stylization artifacts appearing in the original real-time style transfer method. Comparisons with alternative real-time style transfer algorithms, capable of adjusting stylization strength, show that our method reproduces style with more details.
\end{abstract}
\section{Introduction}
Gatys et al.~\cite{gatys2016image} demonstrated that deep neural networks can represent not only the content but also the style of the image which can be described by the Gramm matrices, containing covariances between activations at different channels of the deep convolutional network. The disentanglement of content and style enabled neural style transfer - a technique to render any content image in the style taken from another style image using deep neural network. Since content image already possesses some inherent style, it becomes necessary to specify the amount of style that needs to be transferred from the style image. Original approach of Gatys et al.~\cite{gatys2016image} allowed to do that by adjusting the weights besides content and style components in the target loss function. However, this approach required computationally expensive optimization process in the space of pixel intensities of the stylized image. 

Later works of Ulyanov et al.~\cite{ulyanov2016texture} and Johnson et al.~\cite{johnson2016perceptual} (which we refer to as \emph{real-time style transfer} or \emph{the baseline method} for short), proposed a fast framework for style transfer. In this approach an image transformer network was trained and then any image could be stylized in real-time by passing it through this network. Since transformer was trained using loss function of Gatys et al. consisting of a weighted sum of style and content loss components, a change in desired stylization strength implied a change in the optimization criteria and thus required training a separate transformer network. This incurred not only computational costs for training multiple models, but also storage costs for keeping them on disk. More importantly, this approach suffered from the limitation that stylization with only a discrete set of stylization strengths could be applied - one for each trained transformer network, whereas stylization strength is inherently a continuous feature.

We propose a modification for the style transfer approach of Johnson et al.~\cite{johnson2016perceptual}, which we name \emph{real-time style transfer with strength control}. Proposed algorithm retains the advantage of the original method - namely, it applies style to any image very fast by just passing it through a feed-forward transformer network. However, it additionally gives the possibility to continuously adjust stylization strength at inference time, demonstrated on fig.~\ref{fig:headline}b-\ref{fig:headline}d.  

The proposed architecture yields comparable quantitative results to the baseline algorithm, measured by the total loss value that is minimized for both methods. Qualitatively, for higher stylization strength it gives stylization of comparable quality to the baseline. Interestingly, for smaller stylization strength proposed method gives results of higher quality by alleviating stylization artifacts (shown on fig.~\ref{fig:headline}a) consistently generated by the baseline method. This observation is illustrated on sample images and further supported by the results of the user evaluation study. 

 Qualitative comparison is provided with another modern stylization methods, capable to control stylization strength at inference time. Namely, we compare our method with AdaIn~\cite{huang2017arbitrary} and universal style transfer~\cite{li2017universal}. Results demonstrate that our method reproduces style with significantly more details, while AdaIn and universal style transfer drop much of the style information. 

The paper is organized as follows. Section~\ref{seq:related_work} gives an overview of related methods. Section~\ref{seq:method} describes the proposed method in detail.  Section~\ref{seq:experiments} provides experimental results comparing proposed algorithm with existing real-time style transfer method qualitatively, quantitatively and by means of a user study. It also provides qualitative comparison of our method with two other  methods capable of performing stylization with strength control. Finally, section~\ref{seq:conclusion} concludes.

\section{Related work}\label{seq:related_work}
The task of rendering image in given style, also known as style transfer and non-photorealistic rendering, is a long studied problem in computer vision. Earlier approaches~\cite{gooch2001non,strothotte2002non,rosin2012image} mainly targeted reproduction of specific styles (such as pencil drawings or oil paintings) and used hand-crafted features for that. Later work of Gatys et. al~\cite{gatys2016image} proposed a style transfer algorithm based on deep convolutional neural network VGG~\cite{simonyan2014very}. This algorithm was not tied to specific style. Instead the style was specified by a separate style image. Key discovery of Gatys et al. was to find representation for content and style based on activations inside the  convolutional neural network. Thus content image could produce target content representation and style image could produce target style representation and for any image we could measure its deviation in style and content, captured by content and style losses. Proposed approach was to find an image giving minimal weighted sum of the content and style loss. Stylization strength was possible by adjusting the weighting factor besides the style loss. 

However, algorithm of Gatys et al. required computationally expensive optimization taking several minutes even on modern GPUs. To overcome this issue Ulyanov et al.~\cite{ulyanov2016texture} and Johnson et al.~\cite{johnson2016perceptual} proposed to train a transformer network for fast stylization. The content image there was simply passed through the transformer network for stylization. The network was trained using a weighted sum of content and style loss of Gatys et al. Thus it was tied to specific style and stylization strength fixed in the loss function and  modification of stylization strength at inference time was not possible.

Further research targeted to propose a transformer network architecture capable of applying different styles simultaneously. Main idea was to hold most of architecture fixed and to vary only particular elements depending on the style. Chen et al.~\cite{chen2017stylebank} used separate convolution filter for each style. Dumoulin et al.~\cite{dumoulin2017learned} used different parameters of instance normalization, but these coefficients still needed to be optimized. Ghiasi et al.~\cite{ghiasi2017exploring} introduced separate style prediction network capable of predicting these coefficients without running a separate optimization process. Each set of instance normalization coefficients encoded particular style. By weighting these coefficients a transformer network could yield combinations of respective styles. However, the problem of stylization strength control was not addressed in these works.

In alternative line of research first two moments of the intermediate autoencoder representation of content image were matched to the first two moments of style representation. Huang et al.~\cite{huang2017arbitrary} matched means and variances of activations at a single intermediate layer. Li et al.~\cite{li2017universal} extended this approach by matching means and whole covariance matrix instead. Moreover,  in their approach a content image was passed through a sequence of autoencoders and its representation was adjusted in each of them. Stylization strength was possible in these methods by targeting  a weighted combination of the content and style moments: higher coefficient besides the style moment imposes more style and vice versa.

Generative adversarial networks~\cite{goodfellow2014generative} (GANs) were also successfully applied to style transfer, for instance, in~\cite{zhu2017unpaired,cao2018cari}. The difference to the framework considered in this paper is that GANs retrieve style from multiple style images instead of just one.

\section{Real-time Style transfer With Strength Control}\label{seq:method}

\subsection{Baseline method}
Our real-time style transfer with strength control architecture is built upon the baseline method -- real-time style transfer of Johnson et al.~\cite{johnson2016perceptual} with minor improvements: batch-normalization layers are replaced with instance normalization layers, following~\cite{ulyanov2016instance}, and transposed convolutions replaced by nearest neighbor upsampling and ordinary convolutions to omit checkerboard artifacts, following~\cite{odena2016deconvolution}. Other specifications of layers are not changed, training details are also fully reproduced except that it is empirically found that 80K images are sufficient for convergence of the transformer network.

\subsection{Proposed extension}
Proposed method is built upon the structure of the baseline method. Besides content image $x_c$, transformer network $T_w(x_c,\alpha)$ also accepts stylization strength $\alpha\in \mathbb{R}$ as input. Stylization is performed by passing content image through the transformer:
$$
x = T_w(x_c,\alpha)
$$
Building blocks of the proposed algorithm remain from the baseline, except residual blocks. In~\cite{johnson2016perceptual} traditional residual block $i=1,2,...5$ for input $u$ produces the output is given by the sum of identity transformation $u$ and a non-linear transformation $f_i(u)$. Our modified residual block $i$ outputs $u+\gamma_i f_i(u)$, where 
\begin{equation} \label{eq:renormalization}
\gamma_i = 2\frac{\abs{\alpha \beta_i}}{1+\abs{\alpha \beta_i}}, 
\end{equation}
and $\beta_i$ is a trainable parameter of block $i$. Renormalization~\ref{eq:renormalization} is performed to ensure that for any stylization strength $\alpha$ non-linear result factor $\gamma_i$ stays within a reasonable range $[0,2)$ and higher $\alpha$ increases the impact of the non-linear transformation $f_i(u)$ on the final result. Since $f_i(u)$ is responsible for applying style and identity skip-connection leaves input image as it is, $\gamma_i$ and thus $\alpha$ naturally controls the amount of style added to the input.

\begin{figure}[t]
	\centering
	\includegraphics[width=0.9\textwidth]{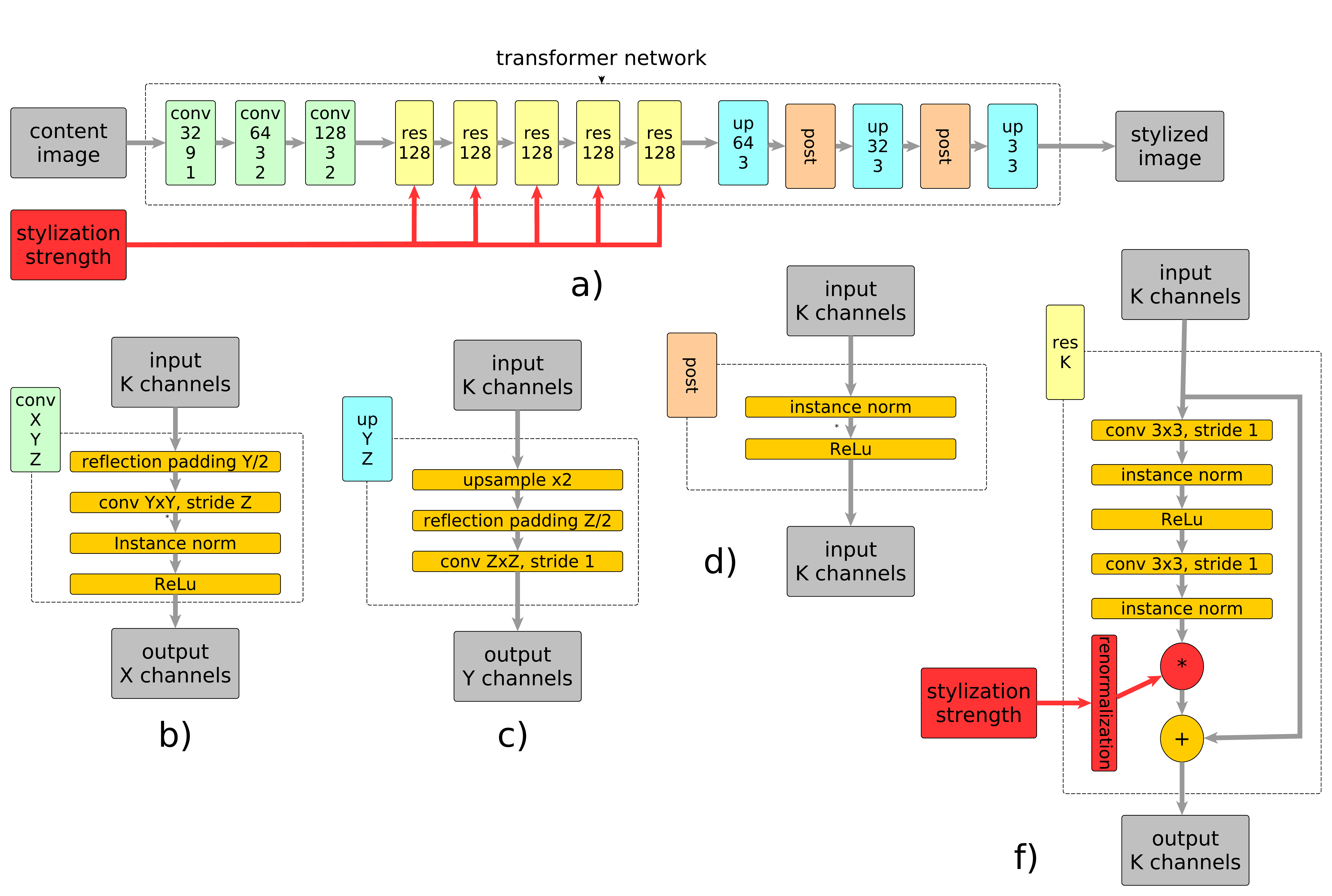}
	\caption{Architecture of real-time style transfer with stylization strength control, differences from the structure of Johnson are highlighted in red. a) transfer network b)-f) detailed schemes of convolutional, residual, upsampling and post-processing blocks respectively.} \label{fig:general_scheme}
\end{figure}

The structure of the proposed method is shown on fig.~\ref{fig:general_scheme} with the new components, compared to the baseline method, highlighted in red.
	
\subsection{Training}
Our method is trained using conventional Gatys loss~\cite{gatys2016image} by randomly sampling content images from MS COCO 2014 training dataset~\cite{zitnick2014microsoft}, consisting of 80K images with a batch size 16. For each mini-batch style strength is sampled uniformly from the grid $[0,0.1,\ldots 10.0]$ and style loss is multiplied by this random factor, thus teaching the generator network to produce images with different stylization strengths. This is an interesting learning paradigm in its own right, when the loss function is modified on each optimization iteration to enforce a range of requirements on the model. Each image is resized and cropped to $256\times256$. One epoch through the dataset is enough for convergence, and training one model takes around 45 minutes on two NVIDIA GeForce GTX 1080 GPUs. We use Adam optimizer with learning rate $10^{-3}$. Total variation strength is set to $10^{-5}$.

\section{Experiments}\label{seq:experiments}

\subsection{Qualitative comparison with real-time style transfer}
\begin{figure}
	\centering
	\includegraphics[width=0.9\textwidth]{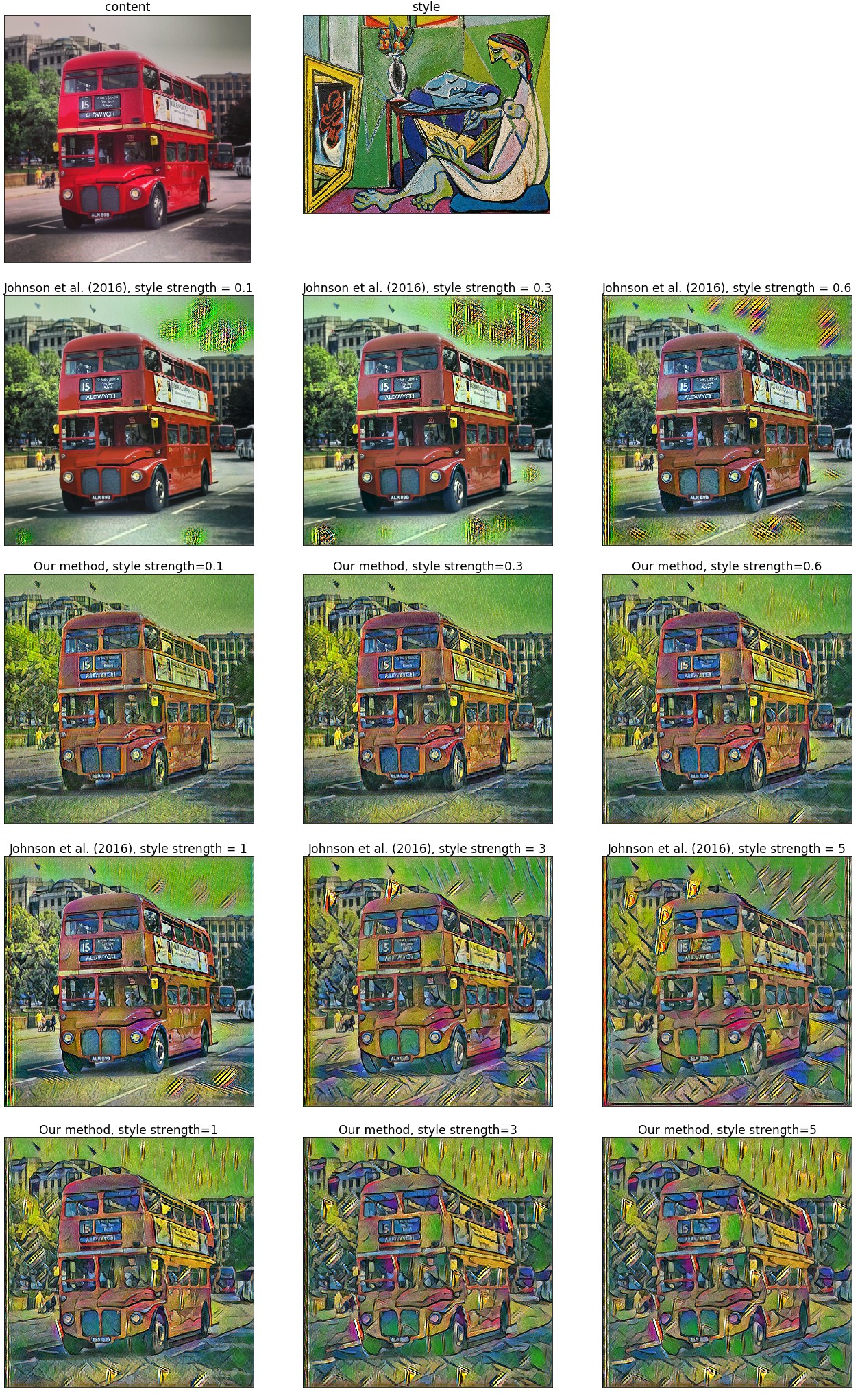}
	\caption{Johnson et al.~\cite{johnson2016perceptual} vs. our method. Proposed algorithm uses single generator network, while baseline needs separate network for each stylization strength. This single generator is enough to stylize with different strength. The method of Johnson et al. frequently generates local artifacts when performing stylization with small strength (second row). Proposed algorithm alleviates these artifacts and produces more pleasing results.} \label{fig:compare_Johnson1}
\end{figure}

\begin{figure}[p]
	\centering
	\includegraphics[width=0.9\textwidth]{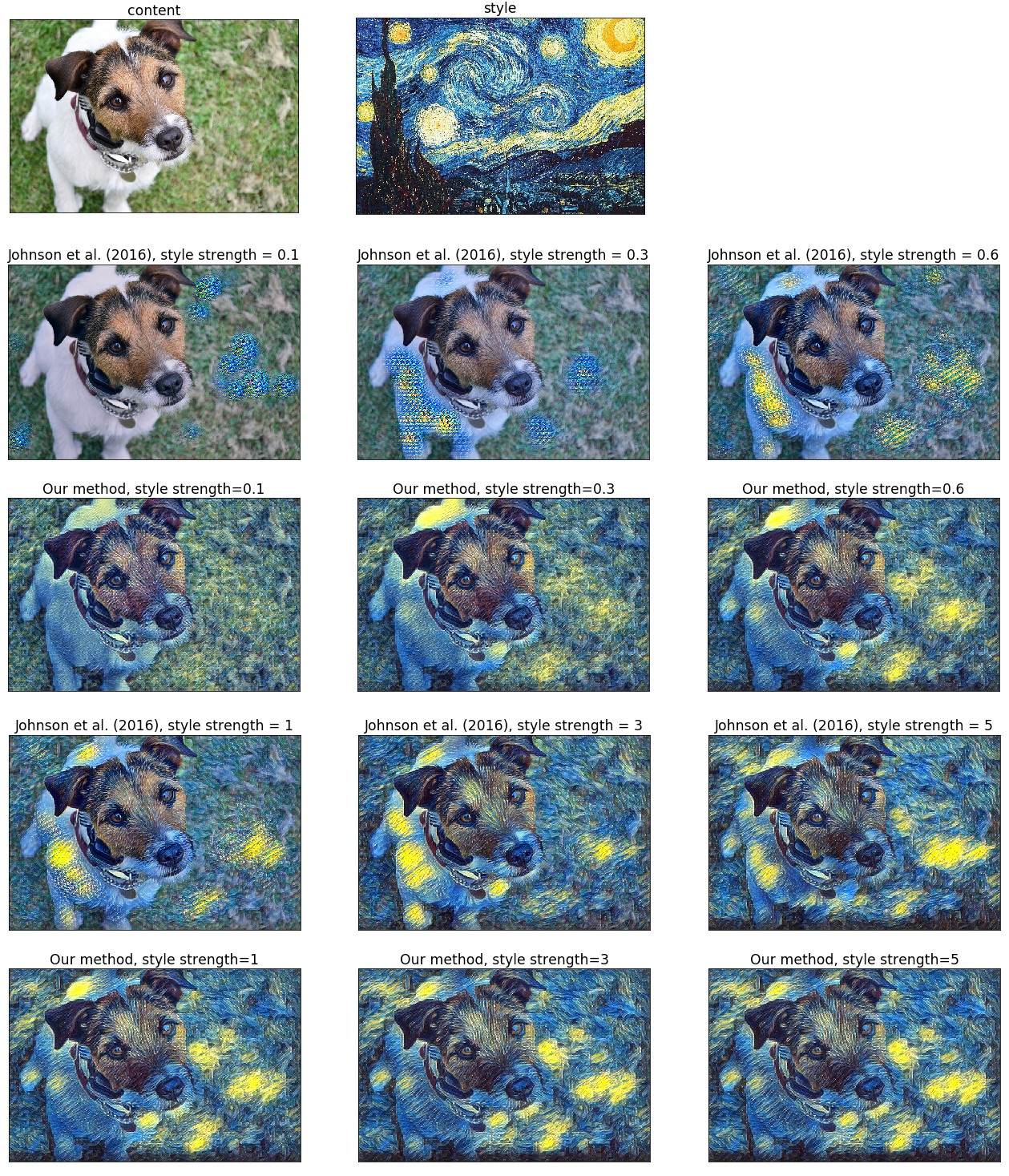}
	\caption{Johnson et al.~\cite{johnson2016perceptual} vs. our method. Proposed algorithm uses single generator network, while baseline needs separate network for each stylization strength. This single generator is enough to stylize with different strength. The method of Johnson et al. frequently generates local artifacts when performing stylization with small strength (second row). Proposed algorithm alleviates these artifacts and produces more pleasing results.} \label{fig:compare_Johnson2}
\end{figure}

A complete implementation of our approach in pytorch with pretrained models is available for download\footnote{https://github.com/Apogentus/style-transfer-with-strength-control}. By applying different styles to various content images in our experiments, it is observed that real-time style transfer of Johnson et al.~\cite{johnson2016perceptual}, being a very powerful model, overfits to the target loss function. This results in sporadically appearing local artifacts when applying generator model trained with small stylization strength. Our model is based upon framework of Johnson et al. but is less flexible due to necessity to apply style with a whole range of different strengths. This additional constraint serves as regularization and reduces overfitting, alleviating observed artifacts during style transfer. It can be seen on fig.~\ref{fig:compare_Johnson1} and fig.~\ref{fig:compare_Johnson2} where our method gives visually more pleasing results without salient artifacts of the baseline model. These findings are consistent for different content and style images.

\subsection{Quantitative comparison with real-time style transfer}
\begin{figure}
	\includegraphics[width=0.9\textwidth]{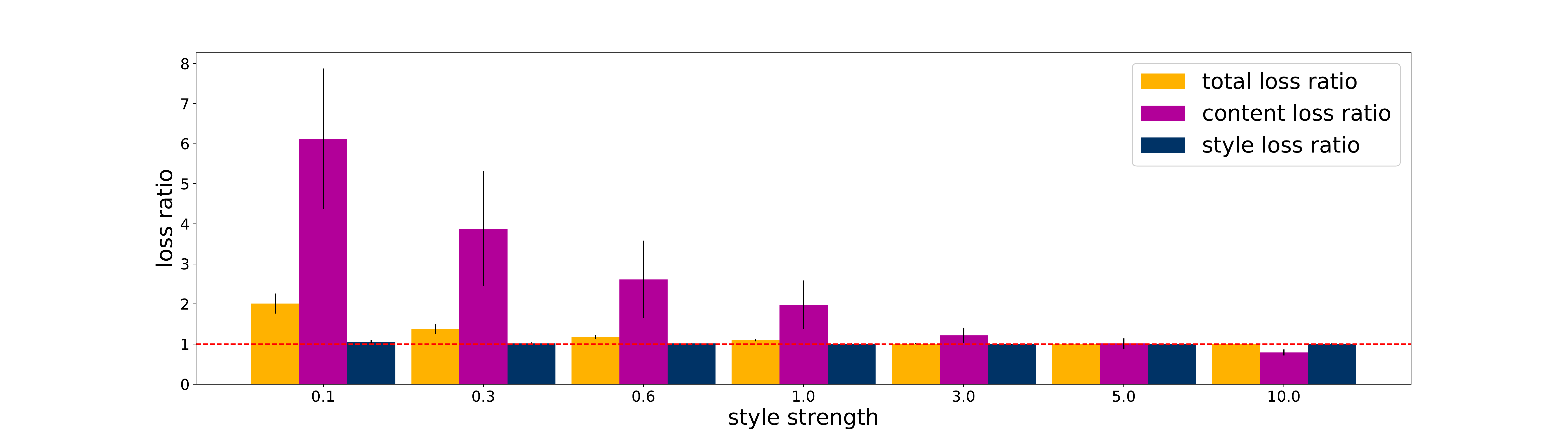}
	\caption{Ratio of our loss function to the Johnson et al. 2016. loss function for total, content and style loss evaluated for different style strengths.} \label{fig:loss_ratios}
\end{figure}

To compare the results quantitatively we consider 5K content images from MS COCO 2017 dataset resized and cropped to $256\times 256$. We stylize these images with the baseline method of Johnson et al.~\cite{johnson2016perceptual} and our method using 8 styles from~\cite{images_sample}. Average style loss is approximately 15 times bigger than average content loss and total variation loss is several orders of magnitude less and thus removed from consideration. We calculate average ratio between our method loss and baseline method loss as well as standard deviation of this ratio along different styles.  Fig.~\ref{fig:loss_ratios} shows ratios for total loss, content loss and style loss. It can be seen that our method closely reproduces total loss - it is close to the baseline total loss for style strength greater or equal to 1 and increases only $2$ times for style strength dropping to $0.1$. Style loss stays very close to the baseline in all cases, so change in the total loss occurs due to significant difference in the content loss. Baseline method indeed is able to achieve lower content loss by preserving content image in all pixels except local regions with strongly expressed style which appears as an artifact and is an undesirable property of the baseline stylizer. 

\subsection{User evaluation study}
Proposed method and the baseline method of Johnson et al.~\cite{johnson2016perceptual} were compared on the representative set of 4 content images and 8 style images~\cite{images_sample}. Each content was stylized using every style with stylization strength randomly chosen from $[0.1,0.3,0.6,1,3,5,10]$, giving a set of 32 stylizations. 9 respondents were sequentially shown pairs of stylizations in the same setup for our and the baseline method. For each pair they were asked to select stylization they like more. To omit position bias, stylizations of the two methods were placed in random order. 

Table~\ref{tab:my_vs_Johnson_user_study} gives a summary of the results. These results suggest that the proposed method gives visually more pleasing results than the baseline method of Johnson et al. in 2/3 of cases. This is an expected result since mentioned above artifacts of the baseline method appear consistently in cases of stylization with small strength.

\begin{table}
	\caption{Summary of the user evaluation study.}\label{tab:my_vs_Johnson_user_study}
	\begin{tabular}{|p{0.91\textwidth}|p{0.09\textwidth}|}
		\hline
	Total number of image pairs &  32\\ \hline
	Total number of respondents &  9\\ \hline
	Total number of responses &  288\\ \hline
	Number of responses when the proposed method was better than the baseline & 192\\ \hline
	The same as proportion & 66.6\%\\ \hline
	Number of images that were better rendered by the proposed method & 21\\ \hline
	The same as proportion & 65.6\%\\
	\hline
\end{tabular}
\end{table}

\begin{figure}[h]
	\centering
	\includegraphics[width=0.9\textwidth]{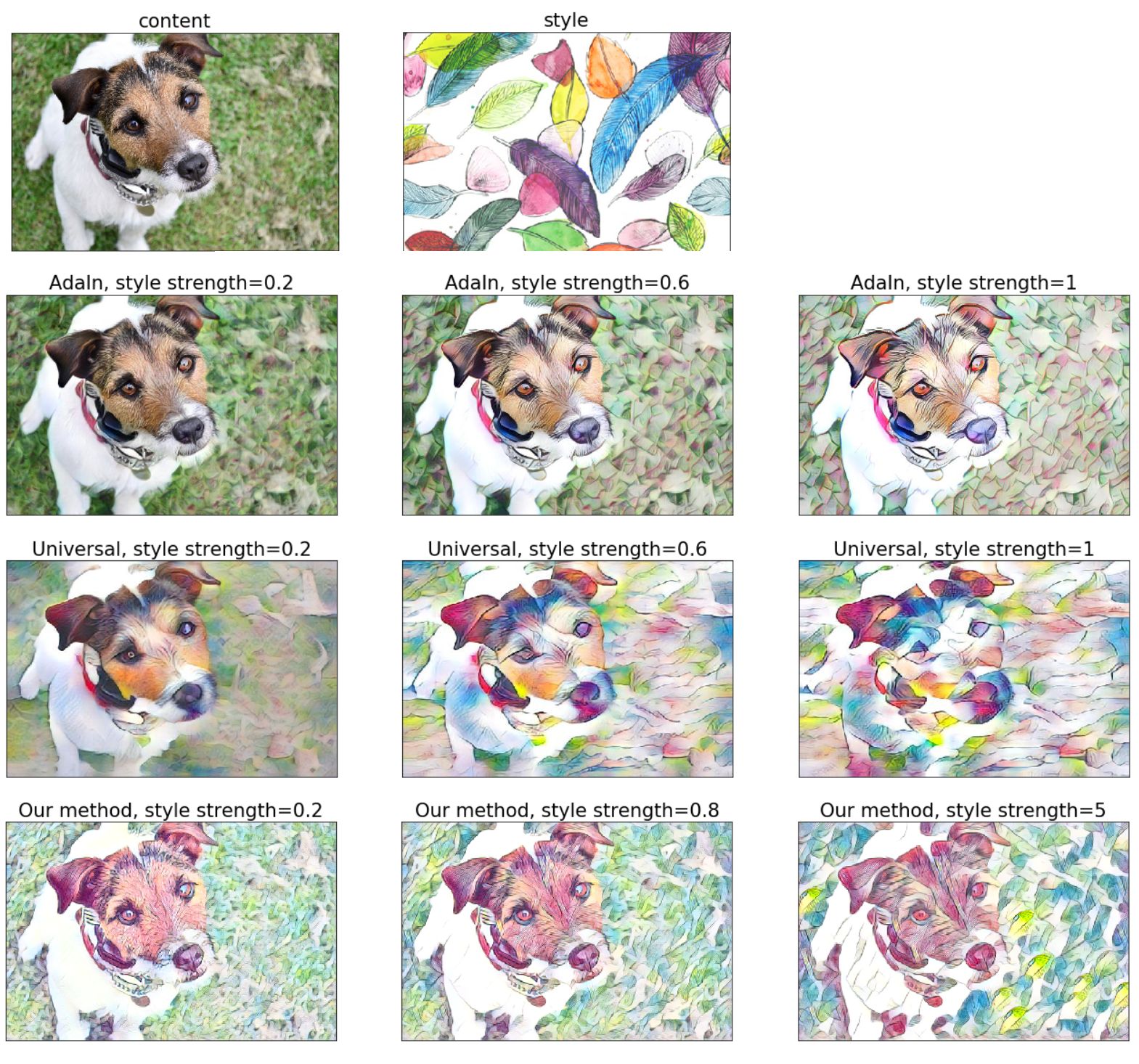}
	\caption{Qualitative comparison of our method with AdaIn and universal style transfer. Our method better reproduces style details.} \label{fig:my_vs_AdaIn_universal_dog}
\end{figure}

\begin{figure}[htbp]
	\centering
	\includegraphics[width=0.85\textwidth]{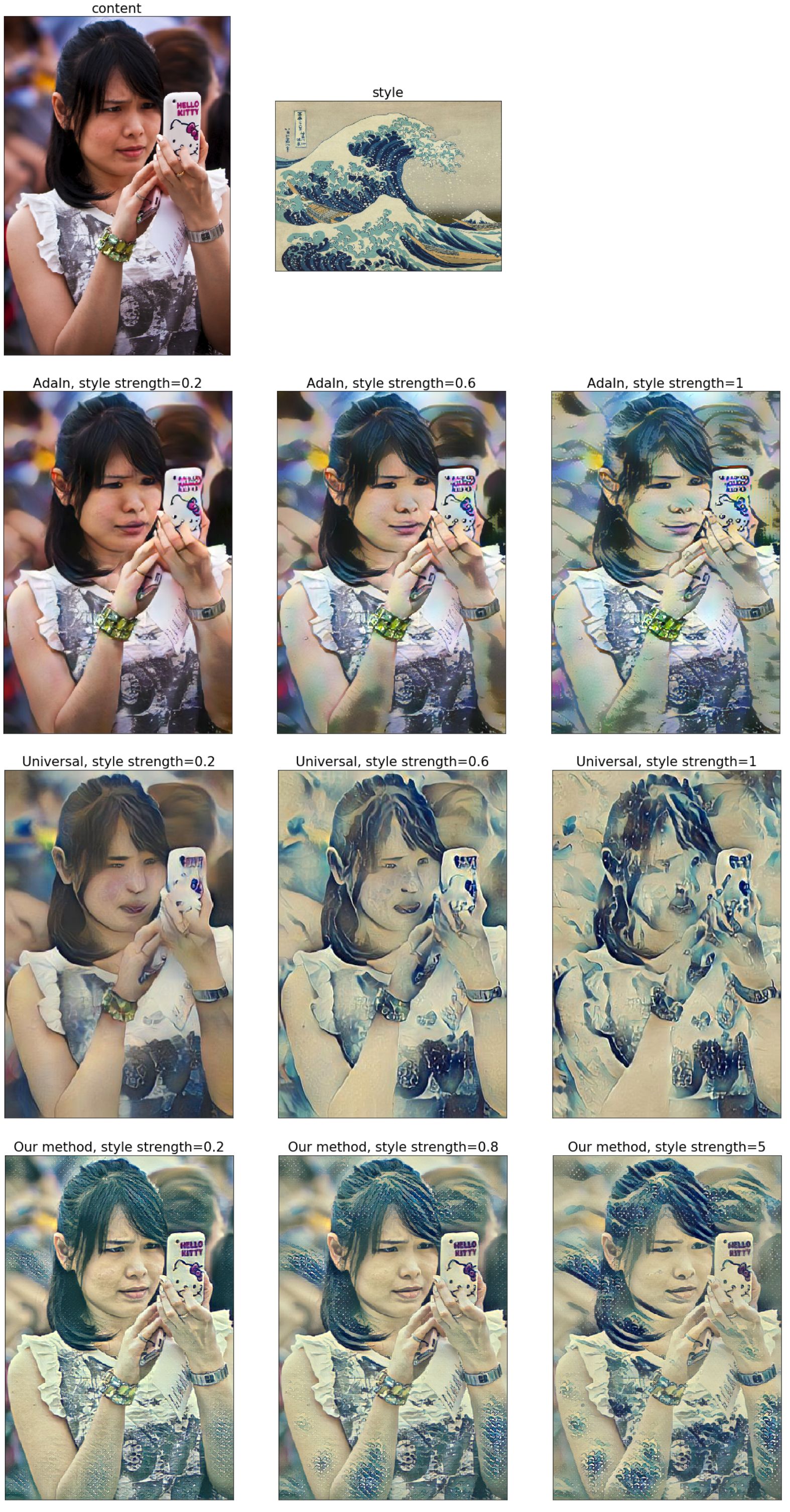}
	\caption{Qualitative comparison of our method with AdaIn and universal style transfer. Our method better reproduces style details.} \label{fig:my_vs_AdaIn_universal_woman_telephone}
\end{figure}

\subsection{Qualitative comparison with other existing methods}
Since AdaIn~\cite{huang2017arbitrary} and universal style transfer~\cite{li2017universal} enable style strength control during inference, we present qualitative comparisons of our and their stylizations on fig.~\ref{fig:my_vs_AdaIn_universal_dog} and fig.~\ref{fig:my_vs_AdaIn_universal_woman_telephone}. Style strength for these methods is controlled by special parameter which is limited to $[0,1]$ interval. So we adapt style strength levels of our method to qualitatively match parameter levels of the compared methods. 

Comparisons on fig.~\ref{fig:my_vs_AdaIn_universal_dog} and fig.~\ref{fig:my_vs_AdaIn_universal_woman_telephone} conform well with the structures of the methods. AdaIn applies style transfer by matching means and standard deviations of the intermediate content image representation to that of the style image. Representation is calculated on single layer of the autoencoder. This simple operation does not allow to reproduce style in detail and generates simplified cartoon-like result instead. Universal style transfer applies style by passing content image through a sequence of autoencoders and adapts mean and whole covariance matrix of the intermediate image representation on each of them. This allows to reproduce more characteristics of style but still without fine details due to limited capabilities of linear scaling. Also since universal style transfer applies style by passing content image through a sequence of deep autoencoders, the result becomes comparatively more blurry. 

Instead of linear scaling our method applies stylization by multiple non-linear transformations which makes it more flexible and allows the method to reproduce style with fine details. Nevertheless, for some applications, such as cartoon-like poster creation, image simplifications obtained by AdaIn and universal style transfer are also desirable properties.

\section{Conclusion}\label{seq:conclusion}
We have presented an extension to the real-time style transfer of Johnson et al.~\cite{johnson2016perceptual} which allows training a single image transformer network capable of stylization with adjustable stylization strength at inference time. Qualitative and quantitative comparisons show that the proposed architecture is good at applying stylization of different strength and produces results not worse than Johnson et al. Although average content loss obtained by their method is lower, it comes at a price of introducing distracting local artifacts to the stylized image. Proposed algorithm alleviates these artifacts, which may be attributed to the regularization effect of the training procedure forcing the model to solve not a particular task, but a range of tasks. Conducted user study supports our qualitative conclusions that the proposed method gives perceptually more appealing stylization results. Qualitative comparisons with other methods capable of real-time stylization strength control show that our algorithm better preserves details of the style. Thus the proposed algorithm is a viable style transfer solution when real-time control of stylization strength is important.

\bibliographystyle{splncs04}
\bibliography{article_bibliography}

\begin{thebibliography}{10}
\providecommand{\url}[1]{\texttt{#1}}
\providecommand{\urlprefix}{URL }
\providecommand{\doi}[1]{https://doi.org/#1}

\bibitem{cao2018cari}
Cao, K., Liao, J., Yuan, L.: Carigans: Unpaired photo-to-caricature translation
  (2018)

\bibitem{chen2017stylebank}
Chen, D., Yuan, L., Liao, J., Yu, N., Hua, G.: Stylebank: An explicit
  representation for neural image style transfer. In: Proceedings of the IEEE
  Conference on Computer Vision and Pattern Recognition. pp. 1897--1906 (2017)

\bibitem{dumoulin2017learned}
Dumoulin, V., Shlens, J., Kudlur, M.: A learned representation for artistic
  style. Proc. of ICLR  \textbf{2} (2017)

\bibitem{gatys2016image}
Gatys, L.A., Ecker, A.S., Bethge, M.: Image style transfer using convolutional
  neural networks. In: Proceedings of the IEEE Conference on Computer Vision
  and Pattern Recognition. pp. 2414--2423 (2016)

\bibitem{ghiasi2017exploring}
Ghiasi, G., Lee, H., Kudlur, M., Dumoulin, V., Shlens, J.: Exploring the
  structure of a real-time, arbitrary neural artistic stylization network.
  arXiv preprint arXiv:1705.06830  (2017)

\bibitem{gooch2001non}
Gooch, B., Gooch, A.: Non-photorealistic rendering. AK Peters/CRC Press (2001)

\bibitem{goodfellow2014generative}
Goodfellow, I., Pouget-Abadie, J., Mirza, M., Xu, B., Warde-Farley, D., Ozair,
  S., Courville, A., Bengio, Y.: Generative adversarial nets. In: Advances in
  neural information processing systems. pp. 2672--2680 (2014)

\bibitem{huang2017arbitrary}
Huang, X., Belongie, S.: Arbitrary style transfer in real-time with adaptive
  instance normalization. In: Proceedings of the IEEE International Conference
  on Computer Vision. pp. 1501--1510 (2017)

\bibitem{johnson2016perceptual}
Johnson, J., Alahi, A., Fei-Fei, L.: Perceptual losses for real-time style
  transfer and super-resolution. In: European conference on computer vision.
  pp. 694--711. Springer (2016)

\bibitem{images_sample}
Kitov, V.: {Set of content and style images}.
  \url{https://github.com/Apogentus/style-transfer-with-strength-control},
  [Online; accessed 1-April-2019]

\bibitem{li2017universal}
Li, Y., Fang, C., Yang, J., Wang, Z., Lu, X., Yang, M.H.: Universal style
  transfer via feature transforms. In: Advances in neural information
  processing systems. pp. 386--396 (2017)

\bibitem{odena2016deconvolution}
Odena, A., Dumoulin, V., Olah, C.: Deconvolution and checkerboard artifacts.
  Distill  (2016). \doi{10.23915/distill.00003},
  \url{http://distill.pub/2016/deconv-checkerboard}

\bibitem{rosin2012image}
Rosin, P., Collomosse, J.: Image and video-based artistic stylisation, vol.~42.
  Springer Science \& Business Media (2012)

\bibitem{simonyan2014very}
Simonyan, K., Zisserman, A.: Very deep convolutional networks for large-scale
  image recognition. arXiv preprint arXiv:1409.1556  (2014)

\bibitem{strothotte2002non}
Strothotte, T., Schlechtweg, S.: Non-photorealistic computer graphics:
  modeling, rendering, and animation. Morgan Kaufmann (2002)

\bibitem{ulyanov2016texture}
Ulyanov, D., Lebedev, V., Vedaldi, A., Lempitsky, V.S.: Texture networks:
  Feed-forward synthesis of textures and stylized images. In: ICML. vol.~1,
  p.~4 (2016)

\bibitem{ulyanov2016instance}
Ulyanov, D., Vedaldi, A., Lempitsky, V.: Instance normalization: The missing
  ingredient for fast stylization. arXiv preprint arXiv:1607.08022  (2016)

\bibitem{zhu2017unpaired}
Zhu, J.Y., Park, T., Isola, P., Efros, A.A.: Unpaired image-to-image
  translation using cycle-consistent adversarial networks. In: Proceedings of
  the IEEE International Conference on Computer Vision. pp. 2223--2232 (2017)

\bibitem{zitnick2014microsoft}
Zitnick, C.L., Dollar, P.: Microsoft coco: Common objects in context. In: ECCV.
  European Conference on Computer Vision (2014)

\end{thebibliography}

\end{document}